\documentclass[runningheads]{llncs}
\usepackage{graphicx}

\usepackage{tikz}
\usepackage{comment}
\usepackage{amsmath,amssymb} 
\usepackage{color}
\usepackage{booktabs}
\usepackage{multirow}
\usepackage{makecell}
\usepackage{threeparttable}
\usepackage{float}
\usepackage{booktabs}
\usepackage{subcaption}
\usepackage{colortbl}
\usepackage[symbol]{footmisc}

\usepackage{xcolor}
\usepackage[colorlinks = true,
            linkcolor = red,
            urlcolor  = magenta,
            citecolor = green,
            anchorcolor = blue]{hyperref}

\let\svthefootnote\thefootnote
\newcommand\blankfootnote[1]{%
  \let\thefootnote\relax\footnotetext{#1}%
  \let\thefootnote\svthefootnote%
}

\usepackage[accsupp]{axessibility}  

\begin{document}
\pagestyle{headings}
\mainmatter
\def\ECCVSubNumber{1545}  

\title{CenterFormer: Center-based Transformer for 3D Object Detection}

\titlerunning{CenterFormer: Center-based Transformer for 3D Object Detection}
\author{Zixiang~Zhou\inst{*1,2} \and
Xiangchen Zhao\inst{1} \and \\
Yu Wang\inst{1} \and Panqu Wang\inst{1} \and Hassan Foroosh\inst{2}}
\authorrunning{Z. Zhou et al.}
\institute{TuSimple \and Computational Imaging Lab., University of Central Florida\\
\email{zhouzixiang@knights.ucf.edu} \email{\{xiangchen.zhao,yu.wang,panqu.wang\}@tusimple.ai}\\
\email{hassan.foroosh@ucf.edu}}

\maketitle

\renewcommand{\thefootnote}{\fnsymbol{footnote}}
\footnotetext[1]{Work done during an internship at TuSimple}

\begin{abstract}
Query-based transformer has shown great potential in constructing long-range attention in many image-domain tasks, but has rarely been considered in LiDAR-based 3D object detection due to the overwhelming size of the point cloud data. In this paper, we propose \textbf{CenterFormer}, a center-based transformer network for 3D object detection. CenterFormer first uses a center heatmap to select center candidates on top of a standard voxel-based point cloud encoder. It then uses the feature of the center candidate as the query embedding in the transformer. To further aggregate features from multiple frames, we design an approach to fuse features through cross-attention. Lastly, regression heads are added to predict the bounding box on the output center feature representation. Our design reduces the convergence difficulty and computational complexity of the transformer structure. The results show significant improvements over the strong baseline of anchor-free object detection networks. CenterFormer achieves state-of-the-art performance for a single model on the Waymo Open Dataset, with 73.7\% mAPH on the validation set and 75.6\% mAPH on the test set, significantly outperforming all previously published CNN and transformer-based methods. Our code is publicly available at \url{https://github.com/TuSimple/centerformer} 
\end{abstract}

\section{Introduction}

\begin{figure}
    \centering
    \includegraphics[width = 0.9\linewidth]{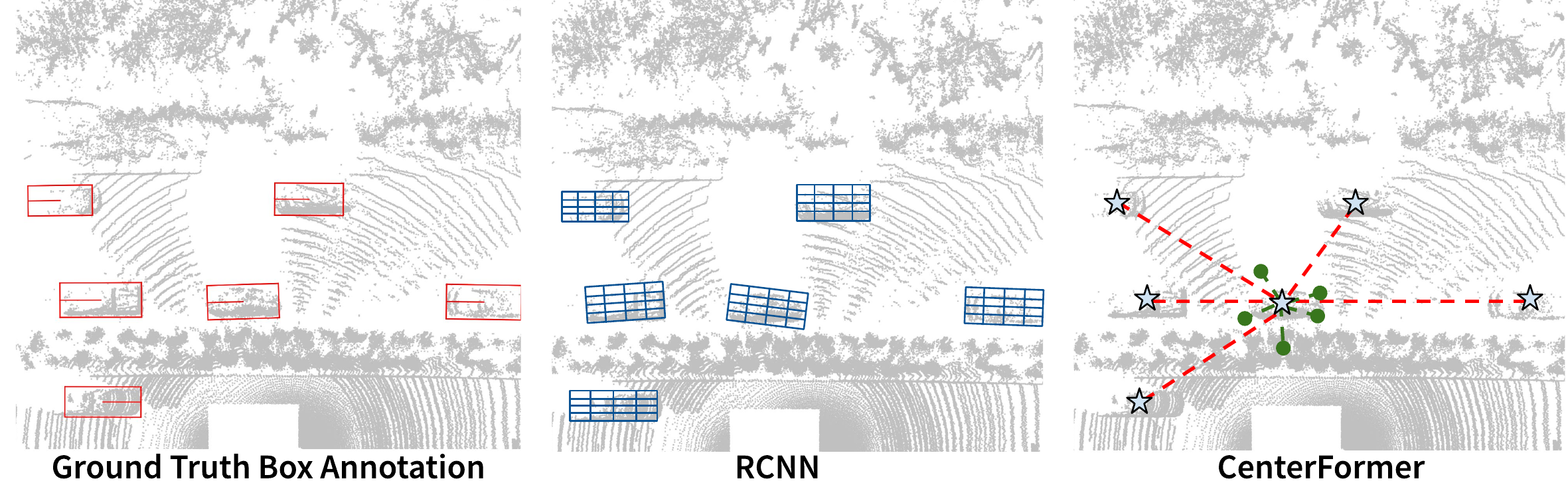}
    \caption{\textbf{The comparison of CenterFormer with RCNN-style detector.} RCNN aggregates point or grid features in RoI, while CenterFormer can learn object-level contextual information and long range features through an attention mechanism.}
    \label{fig:teaser}
\end{figure}

LiDAR is an important sensing and perception tool in autonomous driving due to its ability to provide highly accurate 3D point cloud data of the scanned environment. LiDAR-based 3D object detection aims to detect the bounding boxes of the objects in the LiDAR point cloud. Compared to image-domain object detection, the scanned points in LiDAR data may be sparse and irregularly spaced depending on the distance from the sensor. Most recent methods rely on discretizing the point clouds into voxels \cite{zhou2018voxelnet,yan2018second} or projected bird's eye view~(BEV) feature maps \cite{lang2019pointpillars} to use 2D or 3D convolution networks. Sometimes, it requires a second stage RCNN~\cite{girshick14CVPR}-style refinement network to compensate for the information loss in the voxelization. However, current two-stage networks~\cite{shi2020pv,yin2021center} lack contextual and global information learning. They only use the local features of the proposal~(RoI) to refine the results. The features in other boxes or neighboring positions that could also be beneficial to the refinement are neglected. Moreover, the environment of the autonomous driving scene is not stationary. The local feature learning has more limitations when using a sequence of scans.

In the image domain, transformer encoder-decoder structure has become a competitive method for the detection~\cite{carion2020end,zhu2020deformable} and segmentation~\cite{wang2021max,cheng2021maskformer} tasks. The transformer is able to capture long-range contextual information in the whole feature map and different feature domains. One of the most representative methods is DETR~\cite{carion2020end}, which uses the parametric query to directly learn object information from an encoder-decoder transformer. DETR is trained end-to-end as a set matching problem to avoid any handcrafting processes like non-maximum suppression~(NMS). However, there are two major problems in the DETR-style encoder-decoder transformer network: First, the computational complexity grows quadratically as the input size increases. This limits the transformer to take only low-dimensional features as input which leads to low performance on small objects. Second, the query embedding is learned through the network so that the training is hard to converge. 

Can we design a transformer encoder-decoder network for the LiDAR point cloud in order to better perceive the global connection of point cloud data? Considering the sheer size of LiDAR point cloud data, and the relatively small sizes of objects to be detected, voxel or BEV feature map representations need to be large enough to keep the features for such objects to be separable. As a result, it is impractical to use the transformer encoder structure on the feature map due to the large input size. In addition, if we use a large feature map for the transformer decoder, the query embedding is also difficult to focus on meaningful attention during training. To mitigate these converging problems, one solution is to provide the transformer with a good initial query embedding and confine the attention learning region to a smaller range. In the center-based 3D object detection network \cite{yin2021center}, the feature at the center of an object is used to capture all object information, hence the center feature is a good substitute for the object feature embedding. Multi-scale image pyramid and deformable convolution~\cite{dai17dcn} are two common methods to increase the receptive field of the feature learning without significantly increasing the complexity. Some recent works~\cite{zhu2020deformable,yang2021focal} apply these two methods in the transformer networks.

Taking the aforementioned aspects into consideration, we propose a center-based transformer network, named \textbf{Center Transformer (CenterFormer)}, for 3D object detection. Specifically, we first use a standard voxel-based backbone network to encode the point cloud into a BEV feature representation. Next, we employ a multi-scale center proposal network to convert the feature into different scales and predict the initial center locations. The feature at the proposed center is fed into a transformer decoder as the query embedding. In each transformer module, we use a deformable cross attention layer to efficiently aggregate the features from the multi-scale feature map. The output object representation then regresses to other object properties to create the final object prediction. As shown in Figure~\ref{fig:teaser}, our method can model object-level connection and long-range feature attention. To further explore the ability of the transformer, we also propose a multi-frame design to fuse features from different frames through cross-attention. We test CenterFormer on the large scale Waymo Open Dataset~\cite{sun2020scalability} and the nuScenes dataset~\cite{caesar2020nuscenes}. Our method outperforms the popular center-based 3D object detection networks which are dominant on public benchmarks by a large margin, achieving state-of-the-art performance, with 73.7\% and 75.6\% mAPH on the waymo validation and test sets, respectively. The contributions of our method can be summarised as follows:

\begin{itemize}
    \item We introduce a center-based transformer network for 3D object detection. 
    \item We use the center feature as the initial query embedding to facilitate learning of the transformer.
    \item We propose a multi-scale cross-attention layer to efficiently aggregate neighboring features without significantly increasing the computational complexity.
    \item We propose using the cross-attention transformer to fuse object features from different frames.
    \item Our method outperforms all previously published methods by a large margin, setting a new state-of-the-art performance on the Waymo Open Dataset.
\end{itemize}

\section{Related Work}

\subsection{LiDAR-based 3D Object Detection}
Compared to the well-established point cloud processing networks like PointNet~\cite{qi2017pointnet} and PointNet++~\cite{qi2017pointnet++}, most recent LiDAR detection and segmentation methods~\cite{zhou2018voxelnet,lang2019pointpillars,Zhang2020polarnet,Zhou2021CVPR,zhu2021cylindrical} voxelize the point cloud in a fixed 3D space into a BEV/voxel representation and use conventional 2D/3D convolutional networks to predict the 3D bounding boxes. Other methods~\cite{wu2018squeezeseg,bewley2021range,sun2021rsn,fan2021rangedet} detect the objects on a projected range image. There are also some methods that use hybrid features along with the voxel network~\cite{shi2020pv,Noh_2021_CVPR,ye2020voxel}, and combine multi-view features in the voxel feature representation~\cite{shi2020pv}. VoxelNet~\cite{zhou2018voxelnet} uses a PointNet inside each voxel to encode all points into a voxel feature. This feature encoder later became an essential method in voxel-based point cloud networks. PointPillar~\cite{lang2019pointpillars} proposes the pillar feature encoder to directly encode the point cloud into the BEV feature map so that only 2D convolution is needed in the network.

Similar to image object detection, 3D object detection methods can be divided into anchor-based~\cite{shi2019pointrcnn,shi2020pv,lang2019pointpillars,yan2018second} and anchor-free~\cite{yin2021center,ge2020afdet} methods. Anchor-based methods detect objects through a classification of all predefined object anchors, while anchor-free methods generally consider objects as keypoints and find those keypoints at the local heatmap maxima. Even though anchor-based methods can achieve a good performance, they rely heavily on hyper-parameter tuning. On the other hand, as the anchor-free methods become more prevailing in image-domain tasks, many 3D and LiDAR works have adopted the same design and show a more efficient and competitive performance. Many works~\cite{shi2020pv,li2021lidar,deng2020voxel,mao2021pyramid} also require an RCNN-style second stage refinement. Feature maps for each bounding box proposal are aggregated through RoIAlign or RoIPool. CenterPoint~\cite{yin2021center} detects objects using center heatmap and regresses other bounding box information using center feature representation. 

Most methods directly concatenate points from different frames based on the ego-motion estimation to use the multi-frame information. This assumes the model can align object features from different frames. However, independently moving objects cause misalignment of features across frames. Recent multi-frame methods~\cite{huang2020lstm,yin2020lidar} use an LSTM or a GNN module to fuse the previous state feature with the current feature map. 3D-MAN~\cite{yang20213d} uses a multi-frame alignment and aggregation module to learn the temporal attention of predictions from multiple frames. The feature of each box is generated from the RoI pooling.

\subsection{Vision Transformer}
Originally proposed in the Natural Language Processing (NLP) community, transformer~\cite{vaswani2017attention} is becoming a competitive feature learning module in computer vision. Compared to traditional CNN, the transformer has a bigger receptive field, and feature aggregation is based on the response learned directly from pairwise features. A transformer encoder~\cite{dosovitskiy2020vit,Liu2021SwinTH,yang2021focal} usually serves as a replacement for the convolution layer in the backbone network. Meanwhile, the transformer decoder uses high-level query feature embedding as the input and extracts features from feature encoding through cross-attention, which is more common in detection and segmentation tasks~\cite{carion2020end,zhu2020deformable,SETR,wang2021max}. DETR~\cite{carion2020end} uses a transformer encoder-decoder structure to predict objects from learned query embedding. Deformable DETR~\cite{zhu2020deformable} improves the DETR training through a deformable attention layer. Some recent methods~\cite{liu2021dab,meng2021conditional,zhang2022dino} show that DETR is easier to converge using guidance like anchor boxes.

\subsection{3D Transformer}
An important design in transformer structure is the position embedding due to permutation invariance of the transformer input. However, 3D point clouds already have position information in them, which leads to deviance in the design of 3D transformers. Point transformer~\cite{zhao2021point} proposes a point transformer layer in the PointNet structure, where the position embedding in the transformer is the pairwise point distances. 3DETR~\cite{misra2021end} and \cite{liu2021} use a DETR-style transformer decoder in the point cloud except that the query embedding in the decoder is sampled from Farthest Point Sampling (FPS) and learned through classification. Voxel Transformer~\cite{mao2021voxel} introduces a voxel transformer layer to replace the sparse convolution layer in the voxel-based point cloud backbone network. SST~\cite{fan2021embracing} uses a Single-stride Sparse Transformer as the backbone network to prevent information loss in downsampling of the previous 3D object detector. CT3D~\cite{Sheng2021ICCV} uses a transformer to learn a refinement of the initial prediction from local points. In contrast to the above methods, our CenterFormer tailors the DETR to work on LiDAR point clouds with lower memory usage and faster convergence. Moreover, CenterFormer can learn both object-level self-attentions and local cross-attentions without requiring a first-stage bounding box prediction.

\section{Method}

\subsection{Preliminaries}
\textbf{Center-based 3D Object Detection} is motivated by the recent anchor-free image-domain object detection methods~\cite{law2018cornernet,Duan2019CenterNetKT}. It detects each object as a center keypoint by predicting a heatmap on the BEV feature map.
Given the output of a common voxel point cloud feature encoder $M\in R^{h*w*c}$, where $h$ and $w$ are the BEV map size and $c$ is the feature dimension, center-based LiDAR object detection predicts both a center heatmap $H\in R^{h*w*l}$ and the box regression $B\in R^{h*w*8}$ through two separated heads. Center heatmap $H$ has $l$ channels, one for each object class. In training, the ground truth is generated from the Gaussian heatmap of the annotated box center. Box regression $B$ contains 8 object properties: the grid offset from the predicted grid center to the real box center, the height of the object, the 3D size, and the yaw rotation angle. During the evaluation, it takes the class and regression predictions at the top $N$ highest heatmap scores and uses NMS to predict the final bounding box.

\noindent\textbf{Transformer Decoder} aggregates features from the source representation to each query based on the query-key pairwise attention. Each transformer module consists of three layers: A multi-head self-attention layer, a multi-head cross-attention layer, and a feed-forward layer. In each layer, there is also a skip connection that connects the input and the output features and layer normalization. Let $f^{q}$ and $f^{k}$ be the query feature and key feature. The multi-head attention can be formulated as:
\begin{equation}
    f_{i}^{out} = \sum_{m=1}^{M} W_{m}[\sum_{j\in\Omega_{j}} \sigma(\frac{Q_{i}K_{j}}{\sqrt{d}})\cdot V_{j}]
\end{equation}
\begin{equation} 
    Q_{i} = f^{q}_{i}W_{q}+E^{pos}_{i},K_{j} = f^{k}_{j}W_{k}+E^{pos}_{j},V_{j}=f^{k}_{j}W_{v}
\end{equation}
where $i$ and $j$ are the indices of query feature and source feature respectively, $m$ is the head index, $\Omega_{j}$ is the set of attending key features, $\sigma$ is the softmax function, $d$ is the feature dimension, $E^{pos}$ is the position embedding and $W$ is the learnable weight. In the self-attention layer, the query feature and the key feature come from the same set of query feature embedding, while in the cross-attention layer, the set of key features is the source feature representation.

\begin{figure}[t]
    \centering
    \includegraphics[width = 0.88\linewidth]{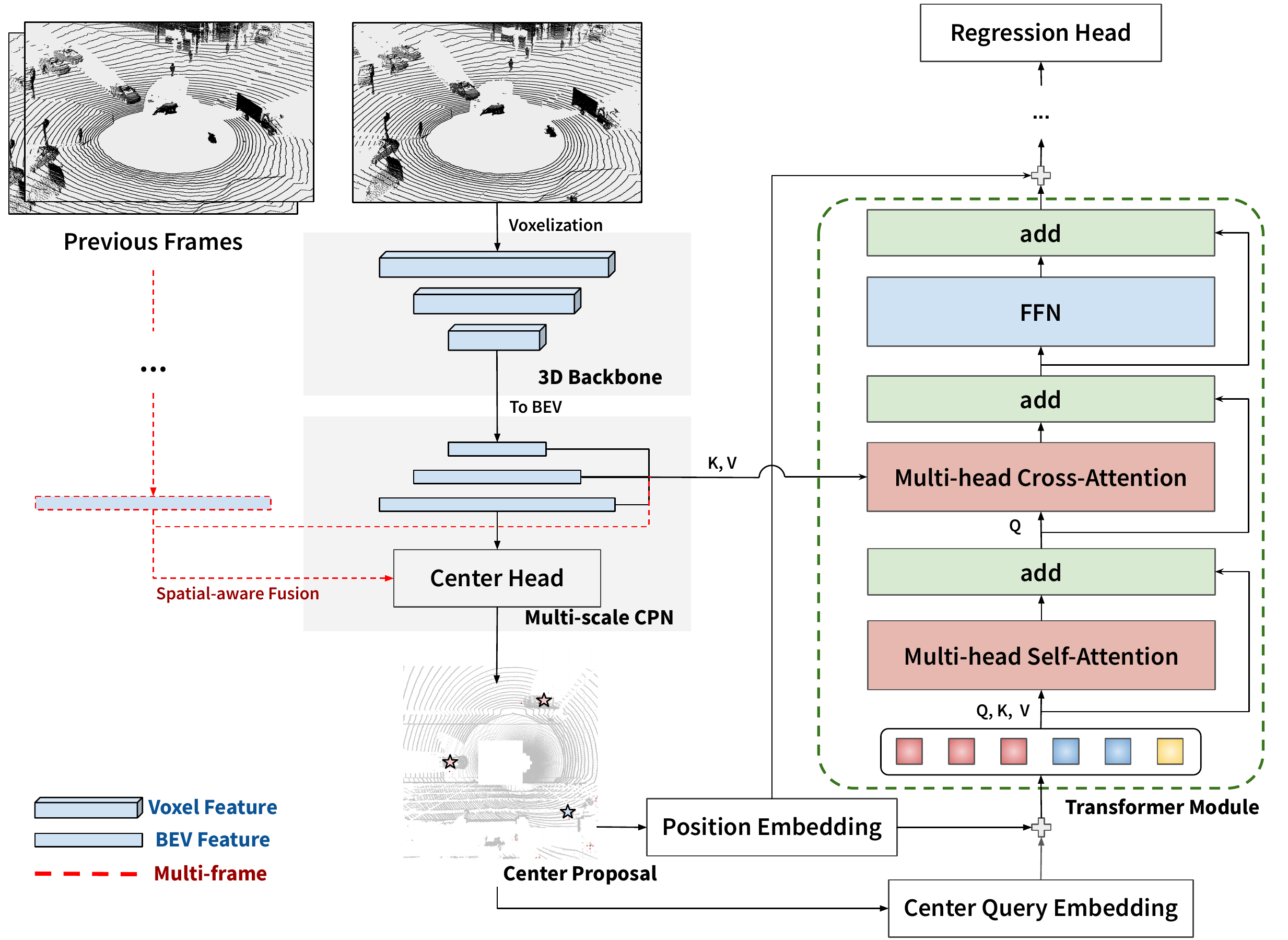}
    \caption{The overall architecture of CenterFormer. The network consists of four parts: a voxel feature encoder that encodes the raw point cloud into a BEV feature representation, a multi-scale center proposal network (CPN), the center-based transformer decoder, and a regression head to predict the bounding box.}
    \label{fig:architecture}
\end{figure}

\subsection{Center Transformer}
The architecture of our model is illustrated in Figure~\ref{fig:architecture}. We use a standard sparse voxel-based backbone network~\cite{yin2021center} to process each point cloud into a BEV feature representation. We then encode the BEV feature into a multi-scale feature map and predict the center proposals. The proposed centers are then used as the query feature embedding in a transformer decoder to aggregate features from other centers and from multi-scale feature maps. Finally, we use a regression head to predict the bounding box at each enhanced center feature. In our multi-frame CenterFormer, the last BEV features of frames are fused together in both the center prediction stage and the cross-attention transformer.

\begin{figure}[t]
\centering
    \begin{subfigure}{0.53\linewidth}
        \centering
        \includegraphics[width=\textwidth]{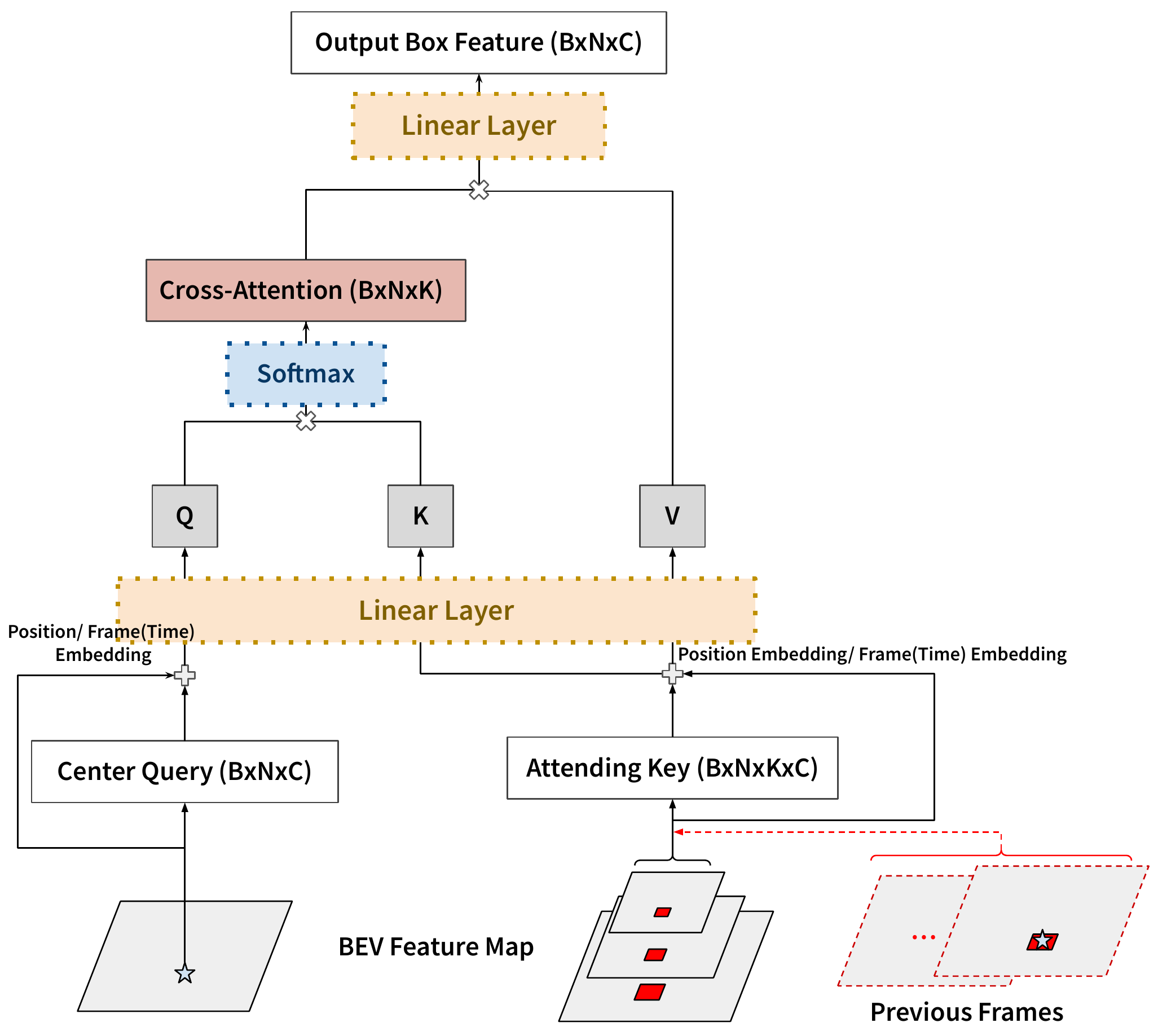}
    \end{subfigure}
    ~
    \begin{subfigure}{0.41\linewidth}
        \centering
        \includegraphics[width=\textwidth]{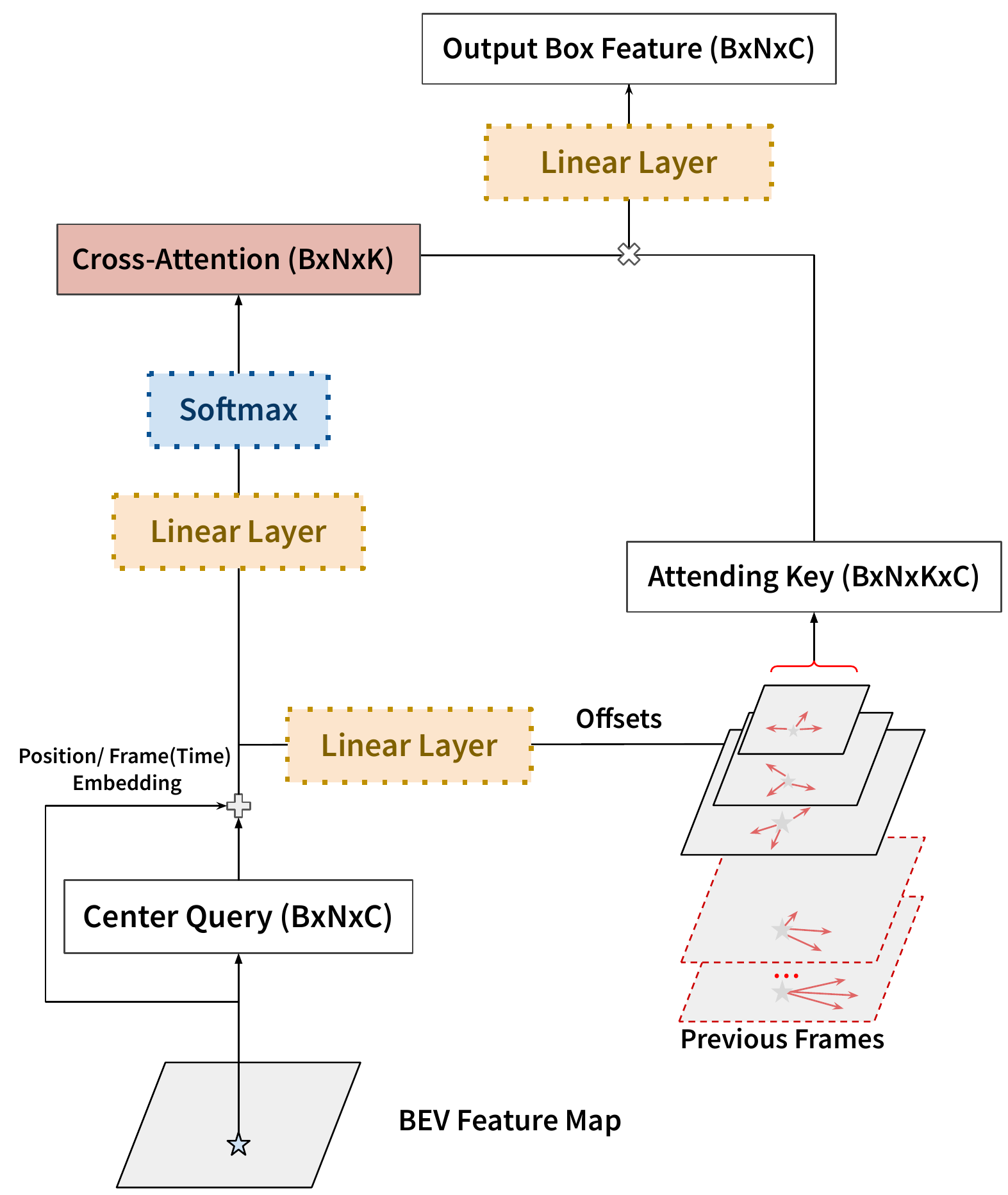}
    \end{subfigure}
\caption{\textbf{Illustration of the cross-attention layer.} (Left) Multi-scale cross-attention. (Right) Multi-scale deformable cross-attention.}
\label{fig:cross_attention}
\end{figure}

\subsubsection{Multi-scale Center Proposal Network}
A DETR-style transformer encoder requires the feature map to be compressed into a small size so that the computation cost is acceptable. This makes the network lose fine-grained features that are crucial for the detection of small objects, which typically occupy $<1\%$ of the space in the BEV map. Therefore, we propose a multi-scale center proposal network (CPN) to replace the transformer encoder for the BEV feature. In order to prepare a multi-scale feature map, we use a feature pyramid network to process the BEV feature representation into three different scales. At the end of each scale, we add a convolutional block attention module (CBAM)~\cite{Woo_2018_ECCV} to enhance the feature via channel-wise and spatial attention.

We use a center head on the highest scale feature map $\mathcal{C}$ to predict an $l$-channel heatmap of object centers. Each channel contains the heatmap score of one class. The location of the top $N$ heatmap scores will be taken out as the center proposals. We used $N=500$ in our experiments empirically.

\subsubsection{Multi-scale Center Transformer Decoder}\label{sssec:msca}
We extract the features at the proposed center locations as the query embedding for the transformer decoder. We use a linear layer to encode the location of the centers into a position embedding. Traditional DETR decoder initializes the query with a learnable parameter. Consequently, the attention weights acquired in the decoder are almost the same among all features. By using the center feature as the initial query embedding, we can guide the training to focus on the feature that contains meaningful object information.  We use the same self-attention layer in the vanilla Transformer decoder to learn contextual attention between objects. The complexity of computing the cross-attention of a center query to all multi-scale BEV features is $O(\sum^{S}_{s=1}h_{s}w_{s}
N)$. Since the BEV map resolution needs to be relatively large to maintain the fine-grained features for small objects, it is impractical to use all BEV features as the attending keypoints. Alternatively, we confine the attending keypoints to a small $3\times3$ window near the center location at each scale, as illustrated in Figure~\ref{fig:cross_attention}. The complexity of this cross-attention is $O(9SN)$, which is more efficient than the normal implementation. Because of multi-scale features, we are able to capture a wide range of features around proposed centers. The Multi-scale cross-attention can be formulated as:

\begin{equation}
    \text{MSCA}(p)=\sum_{m=1}^{M}W_{m}[\sum_{s=1}^{S}\sum_{j\in \Omega_{j}}\sigma(\frac{Q_{i}K^{s}_{j}}{\sqrt{d}})\cdot V^{s}_{j}],
\end{equation}
where $p$ denotes the center proposal, $\Omega_{j}$ here is the window around the center, and $s$ is the index of the scale. The feed-forward layer is also kept unchanged.

\subsubsection{Multi-scale Deformable Cross-attention layer} 
Inspired by~\cite{zhu2020deformable}, we also used a deformable cross-attention layer to sample the attending keypoints automatically. Figure~\ref{fig:cross_attention} shows the structure of the deformable cross-attention layer. Compared to the normal multi-head cross-attention layer, deformable cross-attention uses a linear layer to learn 2D offsets $\Delta p$ of the reference center location $p$ at all heads and scales. The feature at $p+\Delta p$ will be extracted as the cross-attention attending feature through bilinear sampling. We use a linear layer to directly learn the attention scores from the query embedding. Features from multiple scales are aggregated together to form the cross-attention layer output:

\begin{equation}
    \text{MSDCA}(p)=\sum_{m=1}^{M}W_{m}[\sum_{s=1}^{S}\sum_{k=1}^{K}\sigma(W_{msk}\mathcal{C}(p))x^{s}(p+\Delta p_{msk})],
\end{equation}
where $x^{s}$ is the multi-scale BEV feature, $\mathcal{C}(p)$ is the center feature, and $\sigma(W_{msk}\mathcal{C}(p))$ is the attention weight. We used $K=15$ in our experiments.

\subsection{Multi-frame CenterFormer}
Multi-frame is commonly used in 3D detection to improve performance. Current CNN-based detectors cannot effectively fuse features from a fast-moving object, while the transformer structure is more suitable for the fusion due to the attention mechanism. To further explore the potential of CenterFormer, we propose a multi-frame feature fusing method using the cross-attention transformer. As shown in Figure~\ref{fig:architecture}, we process each frame individually using the same backbone network. The last BEV feature of the previous frames is transformed to current coordinates and fused with the current BEV feature in both the center head and cross-attention layer. 

\begin{figure}[t]
    \centering
    \includegraphics[width=0.9\linewidth]{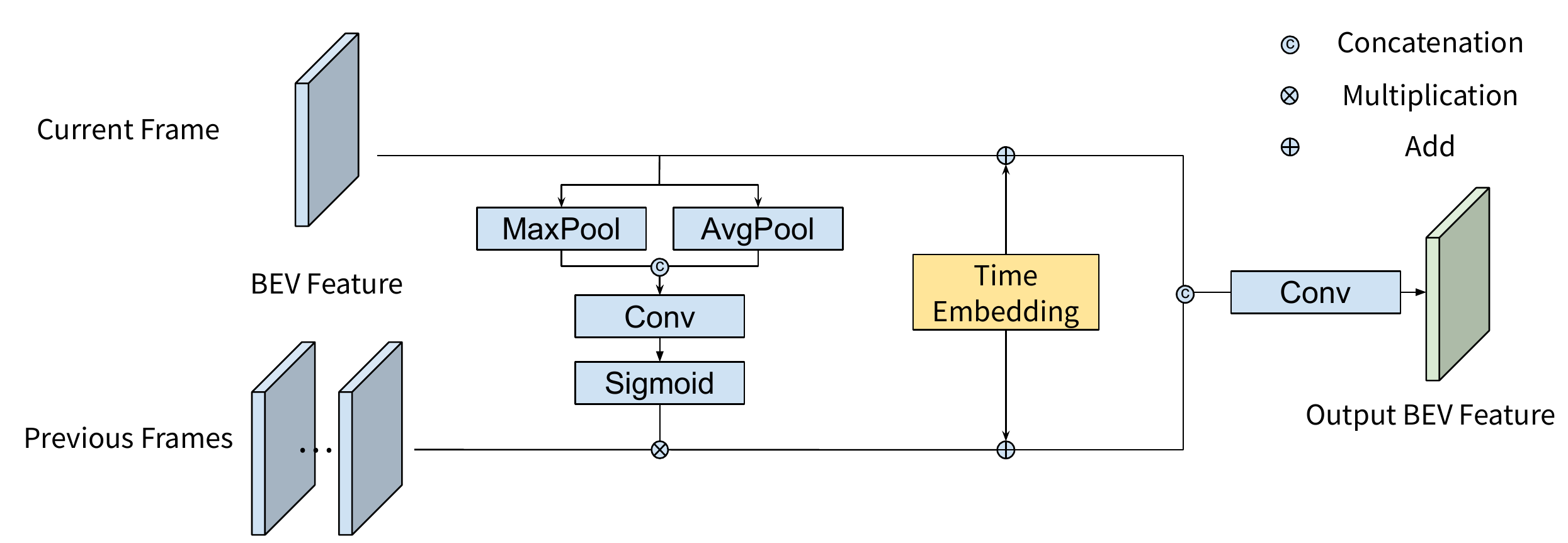}
    \caption{\textbf{The network structure of spatial-aware fusion.} To focus on the current centers, we use current BEV feature as the reference to learn attention.}
    \label{fig:SAF}
\end{figure}

Due to object movements, the center of an object may shift in different frames. Since we only need to predict the center in the current frame, we use a spatial-aware fusion in the center head to alleviate the misalignment error. As shown in Figure~\ref{fig:SAF}, the spatial-aware module uses a similar spatial attention layer as CBAM~\cite{Woo_2018_ECCV} to calculate pixel-wise attention based on the current BEV feature. We concatenate the current BEV feature and weighted previous BEV feature and use an additional convolution layer to fuse them together. We also add the time embedding to the BEV features based on their relative time. Finally, we feed the output fused features to the center head to predict the center candidates.

In the cross-attention layer, we use the location of the center proposal to find the corresponding features in the aligned previous frames. The extracted features will be added to the attending keys. Since our normal cross-attention design uses features in a small window close to the center location, it has limited learnability if the object was out of the window area due to fast movement. Meanwhile, our deformable cross-attention is able to model any level of movement, and is more suitable for the long time-range case. Because our multi-frame model only needs the final BEV feature of the previous frame, it is easy to be deployed to the online prediction by saving the BEV feature in a memory bank.

\subsection{Loss Functions}

Besides the general classification and regression loss functions, we add two additional loss functions to better account for the center-based object detection. First, Inspired by the design in CIA-SSD~\cite{zheng2020ciassd}, we move the IoU-aware confidence rectification module from the second stage of other methods to the regression head. More specifically, we predict an IoU score $iou$ for each bounding box proposal, which is supervised with the highest IoU between the prediction and all ground truth annotations in a smooth L1 loss. During the evaluation, we rectify the confidence score with the predicted IoU score using $\alpha' = \alpha * iou^{\beta}$, where $\alpha$ is the confidence score and $\beta$ is a hyperparameter controlling the degree of rectification. Second, similar to \cite{wang2020centernet3d,hu2022afdetv2}, we also added a corner heatmap head alongside the center heatmap head as auxiliary supervision. For each box, we generate the corner heatmap of four bounding box edge centers and the object center using the same methods to draw the center heatmap except that the Gaussian radius is half size. During training, we supervise the corner prediction with an MSE loss on the region where the ground truth heatmap score is above 0.

The final loss used in our model is the weighted combination of the following four parts: $\mathcal{L} = w_{hm}\mathcal{L}_{hm} + w_{reg}\mathcal{L}_{reg} + w_{iou}\mathcal{L}_{iou} + w_{cor}\mathcal{L}_{cor}.$ We use focal loss~\cite{lin2017focal} and L1 loss for the heatmap classification and box regression. The weights of heatmap classification loss, box regression loss, IoU rectification loss, and corner classification loss are [1, 2, 1, 1], respectively, in our experiment.

\section{Experiments}

We present our experimental results on two large-scale LiDAR object detection benchmarks: Waymo Open Dataset~\cite{sun2020scalability} and nuScenes dataset~\cite{caesar2020nuscenes}. Due to page limitation, the experimental results on the nuScenes dataset, more analysis as well as the details on the choice of network parameters, and additional visualizations are included in the supplementary material.

\subsection{Dataset}

\textbf{Waymo Open Dataset (WOD)} is a large-scale LiDAR point cloud dataset for the autonomous driving environment. It contains 798, 202, and 150 sequences for training, validation, and testing, respectively. Each sequence is 20-second long, captured by a 10 FPS LiDAR sensor with 64 lines in $360^{\circ}$. WOD provides bounding box annotations for three classes: vehicles, pedestrians, and cyclists. The evaluation metrics used in WOD are mean average precision (mAP) and mAP weighted by heading accuracy (mAPH). Each object is categorized into two levels of difficulty, where LEVEL\_1 (L1) denotes that there are more than 5 points on the object and LEVEL\_2 (L2) denotes that there are $1\sim5$ points on the object or the object is manually labeled as L2. The evaluation of the primary metric mAPH L2 includes both L1 and L2 objects. We set the range of the 3D voxel space as $[-75.2\text{m},75.2\text{m}]$ for the $X$ and $Y$ axes, and $[-2\text{m},4\text{m}]$ for the $Z$ axis. The size of each voxel is set to $(0.1\text{m},0.1\text{m},0.15\text{m})$.

\subsection{Implementation Details}
We follow the same VoxelNet backbone network design as~\cite{zhu2019class,yin2021center,shi2020pv}. In our center proposal network, we process the output of the BEV feature into three scales through one upsample layer and one downsample layer. We set the transformer layer/head numbers to 3/4 when using the normal cross-attention, and to 2/6 when using the deformable cross-attention. During evaluation, we use the NMS IoU threshold of $[0.8,0.55,0.55]$ and $\beta=[1,1,4]$ for vehicle, pedestrian and cyclist. For our 8 frames model, we use $\beta=[1,1,1]$ to get a better result for the cyclist.  We also increase the center proposal number $N$ to 1000 in evaluation. We trained our model using AdamW optimizer with the one-cycle policy. We trained the single-frame and multi-frame models on 8 Nvidia A100 GPUs with batch sizes 32 and 16. Due to the memory limitation, 4 frames and 8 frames are first split into two 2 frames and 4 frames mini-batch. The points from frames in each mini-batch will be first concatenated together. Hence our multi-frame model only needs to fuse two BEV features together. We apply the object copy \& paste augmentation during training with the same fade strategy in \cite{wang2021pointaugmenting}. More details are included in the supplementary material.

\subsection{Object Detection Results}
\begin{table}[t]
\centering
\caption{The detection results on WOD validation set. $\dagger$: Deformable CenterFormer. $\ddagger$ Reported by PVRCNN++.}
\label{table:validation}
\resizebox{\linewidth}{!}{
\begin{tabular}{l|c|cc|cc|cc|>{\columncolor[gray]{0.95}}c }
\Xhline{4\arrayrulewidth}
\multicolumn{1}{c|}{Method} & Frame & \begin{tabular}[c]{@{}c@{}}Vehicle L1\\ (mAP/APH)\end{tabular} & \begin{tabular}[c]{@{}c@{}}Vehicle L2\\ (mAP/APH)\end{tabular} & \begin{tabular}[c]{@{}c@{}}Pedestrain L1\\ (mAP/APH)\end{tabular} & \begin{tabular}[c]{@{}c@{}}Pedestrain L2\\ (mAP/APH)\end{tabular} & \begin{tabular}[c]{@{}c@{}}Cyclist L1\\ (mAP/APH)\end{tabular} & \begin{tabular}[c]{@{}c@{}}Cyclist L2\\ (mAP/APH)\end{tabular} & \begin{tabular}[c]{@{}c@{}}Mean L2\\ mAPH\end{tabular} \\ \Xhline{4\arrayrulewidth}
StarNet~\cite{ngiam2019starnet}& 1     & 55.1/54.6 & 48.7/48.3  & 68.3/60.9  & 59.3/52.8 & - / -  & - / -  & -  \\
SECOND$^{\ddagger}$~\cite{yan2018second}& 1     & 72.3/71.7 & 63.9/63.3  & 68.7/58.2  & 60.7/51.3 & 60.6/59.3  & 58.3/57.1  & 57.2  \\
LiDAR R-CNN~\cite{li2021lidar}  & 1     & 73.5/73.0 & 64.7/64.2  & 71.2/58.7  & 63.1/51.7 & 68.6/66.9  & 66.1/64.4  & 60.1  \\
Part-A2$^{\ddagger}$~\cite{shi2020points} & 1     & 77.1/76.5 & 68.5/68.0  & 75.2/66.9  & 66.2/58.6 & 68.6/67.4 &  66.1/64.9  & 63.8  \\
3D-MAN~\cite{yang20213d}& 16     & 74.5/74.0 & 67.6/67.1  & 71.7/67.7  & 62.6/59.0 & - / -  & - / -  & -  \\
PV-RCNN++~\cite{shi2022pv}       & 1     & 79.1/78.6 & 70.3/69.9  & 80.6/74.6  & 71.9/66.3 & 73.5/72.4 &  70.7/69.6  & 68.6  \\
CenterPoint~\cite{yin2021center}& 1     &   - / -   & - /67.9    & - / -      & - /65.6     & - / -   &   - /68.6     & 67.4  \\
CenterPoint~\cite{yin2021center}& 2     &   - / -   & - /69.7      & - / -    & - /70.3    &  - / -   &   - /70.9     & 70.3  \\\Xhline{4\arrayrulewidth}
CenterFormer                    & 1     & 75.0/74.4 & 69.9/69.4  & 78.0/72.4  & 73.1/67.7 & 73.8/72.7  & 71.3/70.2  & 69.1  \\
CenterFormer$^{\dagger}$          & 1     & 75.2/74.7 & 70.2/69.7  & 78.6/73.0  & 73.6/68.3 & 72.3/71.3  & 69.8/68.8  & 69.0  \\
CenterFormer                    & 2     & 77.1/76.6 & 72.2/71.7  & 80.9/77.6  & 76.2/73.0 & 76.0/75.1  & 73.6/72.7  & 72.5  \\
CenterFormer$^{\dagger}$          & 2     & 77.0/76.5 & 72.1/71.6  & 81.4/78.0  & 76.7/73.4 & \textbf{76.6}/\textbf{75.7} &  \textbf{74.2}/\textbf{73.3}  & 72.8  \\
CenterFormer$^{\dagger}$          & 4     & 78.1/77.6 & 73.4/72.9  & 81.7/78.6  & 77.2/74.2 & 75.6/74.8 &  73.4/72.6  & 73.2  \\
CenterFormer$^{\dagger}$          & 8     & \textbf{78.8}/\textbf{78.3} & \textbf{74.3}/\textbf{73.8}  & \textbf{82.1}/\textbf{79.3}  & \textbf{77.8}/\textbf{75.0} & 75.2/74.4 &  73.2/72.3  & \textbf{73.7}
\\\Xhline{4\arrayrulewidth}                                                
\end{tabular}
}
\end{table}

In Table~\ref{table:validation}, we show the results on the validation set. Anchor-free center-based method CenterPoint achieves better performance than the anchor-based methods. PVRCNN++ is the best anchor-based method so far, yet shows weak performance on small objects. This demonstrates the limitation of using manually designed anchors for the detection of objects that have large size variations. Our single frame model outperforms CenterPoint by 1.7\%. Using a multi-frame model can significantly increase the mAPH performance. Our 2/4/8 frames model reaches the mAPH of 72.8\%, 73.2\% and 73.7\%, respectively, becoming the new state-of-the-art. The pedestrian class benefits most from the multi-frame since the pedestrian point cloud suffers most from the occlusion and noise, as well as its small size. The overall better performance verifies the effectiveness of our proposed transformer model.

\begin{table}[t]
\centering
\caption{The single-model detection result on WOD testing set.}
\label{table:testing}
\resizebox{\linewidth}{!}{
\begin{tabular}{l|cc|cc|cc|>{\columncolor[gray]{0.95}}c }
\Xhline{4\arrayrulewidth}
\multicolumn{1}{c|}{Method}  & \begin{tabular}[c]{@{}c@{}}Vehicle L1\\ (mAP/APH)\end{tabular} & \begin{tabular}[c]{@{}c@{}}Vehicle L2\\ (mAP/APH)\end{tabular} & \begin{tabular}[c]{@{}c@{}}Pedestrain L1\\ (mAP/APH)\end{tabular} & \begin{tabular}[c]{@{}c@{}}Pedestrain L2\\ (mAP/APH)\end{tabular} & \begin{tabular}[c]{@{}c@{}}Cyclist L1\\ (mAP/APH)\end{tabular} & \begin{tabular}[c]{@{}c@{}}Cyclist L2\\ (mAP/APH)\end{tabular} & \begin{tabular}[c]{@{}c@{}}Mean L2\\ mAPH\end{tabular} \\ \Xhline{4\arrayrulewidth}
StarNet~\cite{ngiam2019starnet} & 61.5/61.0 & 54.9/54.5  & 67.8/59.9  & 61.1/54.0 & - / -  & - / -  & -  \\
PPBA~\cite{cheng2020improving}  & 64.6/64.1 & 56.2/55.8  & 69.7/61.7  & 63.0/55.8 & - / -  & - / -  & -  \\
M3DETR~\cite{Guan_2022_WACV}    & 77.7/77.1 & 70.5/70.0  & 68.2/58.5  & 60.6/52.0 & 67.3/65.7 &  65.3/63.8  & 61.9  \\
3D-MAN~\cite{yang20213d}        & 78.7/78.3 & 70.4/70.0  & 70.0/66.0  & 64.0/60.3 & - / - &  - / -  &  -  \\
RSN~\cite{bewley2021range}       & 80.7/80.3 & 71.9/71.6  & 78.9/75.6  & 70.7/67.8 & - / - &  - / -  &  -  \\
PV-RCNN++~\cite{shi2022pv}       & 81.6/81.2 & 73.9/73.5  & 80.4/75.0  & 74.1/69.0 & 71.9/70.8 &  69.3/68.2  & 70.2  \\
CenterPoint~\cite{yin2021center}& 81.1/80.6 & 73.4/73.0  & 80.5/77.3  & 74.6/71.5 & 74.6/73.7 &  72.2/71.3  & 71.9  \\
SST\_3f~\cite{fan2021embracing}     & 81.0/80.6 & 73.1/72.7  & 83.3/79.7  & 76.9/73.5 & 75.7/74.6 &  73.2/72.2  & 72.8  \\
AFDetV2~\cite{hu2022afdetv2}     & 81.7/81.2 & 74.3/73.9  & 81.3/78.1  & 75.5/72.4 & \textbf{76.4}/\textbf{75.4} &   \textbf{74.1}/\textbf{73.0}  & 73.1  \\
\Xhline{4\arrayrulewidth}
CenterFormer                    & \textbf{84.7}/\textbf{84.4} & \textbf{78.1}/\textbf{77.7} &  \textbf{84.6}/\textbf{81.8} &  \textbf{79.4}/\textbf{76.6} & 75.5/74.5 &  73.3/72.4  & \textbf{75.6} \\\Xhline{4\arrayrulewidth}                                                
\end{tabular}
}
\end{table}

In Table~\ref{table:testing}, we show the single model results on the test set. The prediction result is submitted to the online server for evaluation. Our method outperforms all the previous methods by a large margin. The result on vehicle and pedestrian classes have significant improvements ($+3.8\%$ and $+3.1\%$ on L2 mAPH) as a result of the long-range contextual information learning of the transformer. 

\begin{table}[t]
\centering
\caption{The comparison of single-frame CenterFormer (trained on vehicle only) with other methods on WOD validation set for vehicle class.}
\label{table:vehicle_only}
\resizebox{0.6\linewidth}{!}{
\begin{tabular}{c|c|c|cc}
\Xhline{4\arrayrulewidth}
&\multicolumn{1}{c|}{Method} &\multicolumn{1}{c|}{Ref} & \multicolumn{1}{c}{L2 mAP} & \multicolumn{1}{c}{L2 mAPH} \\ \Xhline{2\arrayrulewidth}
\multirow{5}{*}{{\rotatebox[origin=c]{90}{CNN}}}
&RSN~\cite{bewley2021range}   & CVPR 2021 &    66.0  &  65.5   \\ 
&Voxel R-CNN~\cite{mao2021voxel} & AAAI 2021 & 66.6 & - \\
&Pyramid R-CNN~\cite{mao2021pyramid} & ICCV 2021 & 67.2 & 66.7 \\
&LiDAR R-CNN~\cite{li2021lidar} & CVPR 2021 &  68.3 & 67.9 \\
&BtcDet~\cite{xu2021behind}   & AAAI 2022 &     70.1      &  69.6    \\ 
\Xhline{2\arrayrulewidth}
\multirow{7}{*}{{\rotatebox[origin=c]{90}{Transformer}}}
&BoxeR-3D~\cite{nguyen2021boxer} & CVPR 2022 & 63.9 & 63.7 \\
&Voxel transformer~\cite{mao2021voxel} & ICCV 2021 & 65.9 &  65.3   \\ 
&M3DETR~\cite{Guan_2022_WACV} & WACV 2022 & 66.6 &  66.0   \\ 
&3D-MAN~\cite{yang20213d} & CVPR 2021 &   67.6 &  67.1   \\ 
&SST\_1f~\cite{fan2021embracing}   & CVPR 2022 &   68.0 &  67.6   \\
&CT3D~\cite{Sheng2021ICCV}  & ICCV 2021  &   69.0     &  -   \\ 
\cline{2-5}
&CenterFormer  & ECCV 2022 &   \textbf{70.7}      &  \textbf{70.2}    \\ 
\Xhline{4\arrayrulewidth}
\end{tabular}
}
\end{table}

To fairly compare with more recent methods~\cite{mao2021voxel,deng2020voxel,mao2021pyramid,Sheng2021ICCV} that train the model and report the result on only the vehicle class, we train the single frame CenterFormer only on Vehicle class too. We show the results in Table~\ref{table:vehicle_only}. As shown in the table, even with the simplest design, CenterFormer outperforms both the recent transformer-based methods and CNN-based baselines.

\subsection{Ablation Study}

\begin{table}[t]
\centering
\caption{The ablation of the LEVEL\_2 mAPH result improvement of each component on the validation set using single frame. SA, CA and DCA denote the self-attention layer, cross-attention layer and deformable cross-attention layer. IoU and Corner denote IoU rectification and Corner auxiliary supervision. Fade denotes the fade augmentation strategy.}
\label{table:ablation}
\resizebox{0.7\linewidth}{!}{
\begin{tabular}{ccccccc|ccc>{\columncolor[gray]{0.95}}c}
\Xhline{4\arrayrulewidth}
CPN & SA & CA & Corner & DCA & IoU & Fade & Vehicle & Pedstrain & Cyclist & Mean \\ \Xhline{2\arrayrulewidth}
 & & & & & & &  66.4                            &     63.4                           &  67.8                            &  65.9    \\
$\checkmark$ & & & & & & &      68.5                        &     62.7                &  67.0                            &  66.1    \\ 
$\checkmark$ & $\checkmark$ & & & & & &      68.7             &     64.5              &  67.3                            &   66.8   \\ 
$\checkmark$ & & $\checkmark$ & & & & &         68.7           &     64.2             &  66.7                            &  66.5    \\ 
$\checkmark$ & $\checkmark$ & $\checkmark$ & & & & &         69.3        &     64.8            &   66.8     &  67.0    \\ \Xhline{2\arrayrulewidth}
$\checkmark$ & $\checkmark$ & $\checkmark$ & $\checkmark$ & & & &         69.5                     &     65.3                           &   66.4                           &  67.1    \\ 
$\checkmark$ & $\checkmark$ & $\checkmark$  & $\checkmark$ &  & $\checkmark$ & &           69.5                     &      66.9                          &    69.7                          &  68.7\\
$\checkmark$ & $\checkmark$ & $\checkmark$ & $\checkmark$ & & $\checkmark$ & $\checkmark$ &         69.4                     &      67.7                          &    70.2                          &  69.1 \\\Xhline{2\arrayrulewidth}
$\checkmark$ & $\checkmark$ & & $\checkmark$ & $\checkmark$ & & &         69.3                     &  65.1                              &   67.5                           &  67.3    \\ 
$\checkmark$ & $\checkmark$ &  & $\checkmark$ & $\checkmark$ & $\checkmark$ &  &           69.2                     &      66.7                          &    69.1                          &  68.3\\

$\checkmark$ & $\checkmark$ &  & $\checkmark$ & $\checkmark$ & $\checkmark$ & $\checkmark$ &           69.7                     &      68.3                          &    68.8                          &  69.0

  \\\Xhline{4\arrayrulewidth}
\end{tabular}
}
\end{table}

In Table~\ref{table:ablation}, we investigate the effect of each added component in our method on a single frame. We use the previous center-based method as the baseline. After changing the RPN to multi-scale CPN and separating the detection into the center proposal and box regression, our method reaches a similar performance despite we flatten the regression head to 1D. The transformer self-attention layer and cross-attention layer both can improve the results, and when used together, the result reaches 67.0\%. This indicates that the self-attention layer and the cross-attention layer learn features separately. Corner auxiliary supervision can additionally improve the result by 0.1\%. On the other hand, deformable cross-attention achieves a better result of 67.3\%. When trained with IoU rectification, the results get a significant boost to 68.7\% and 68.3\% for the models using cross-attention and deformable cross-attention. The fade augmentation strategy can further improve the result by 0.4\% and 0.7\%. This is because the model can adjust to the real data distribution at the end of the training.

\subsection{Analysis}

\textbf{Comparison with deformable DETR}
Deformable DETR~\cite{zhu2020deformable} aims to speed up the learning speed and reduce the computation cost in the DETR structure. Compared to Deformable DETR, our method has three major differences. First, we completely remove the transformer encoder to enable a larger encoded feature map. Second, we use the center feature rather than the learnable parameter as the query embedding for the transformer decoder. Experiment shows that using the center feature as the initial query embedding outperforms the parametric embedding by 1.5\% mAPH. This is because the center feature already contains object-level information, which makes it easier to learn pairwise attentions. Third, we use a similar training strategy as \cite{yin2021center} rather than the end-to-end set matching training strategy. The set matching training is known for being hard to converge. Since we already have an initial center proposal, we can limit the network to learn only when the proposal is close to the ground truth annotation to speed up the training. Experiment shows that if we use the same set matching training strategy in DETR, the mAPH result is 46.3\%, which is more than 20\% lower than our current training method.

\begin{figure}[t]
    \centering
    \includegraphics[width = 0.85\linewidth]{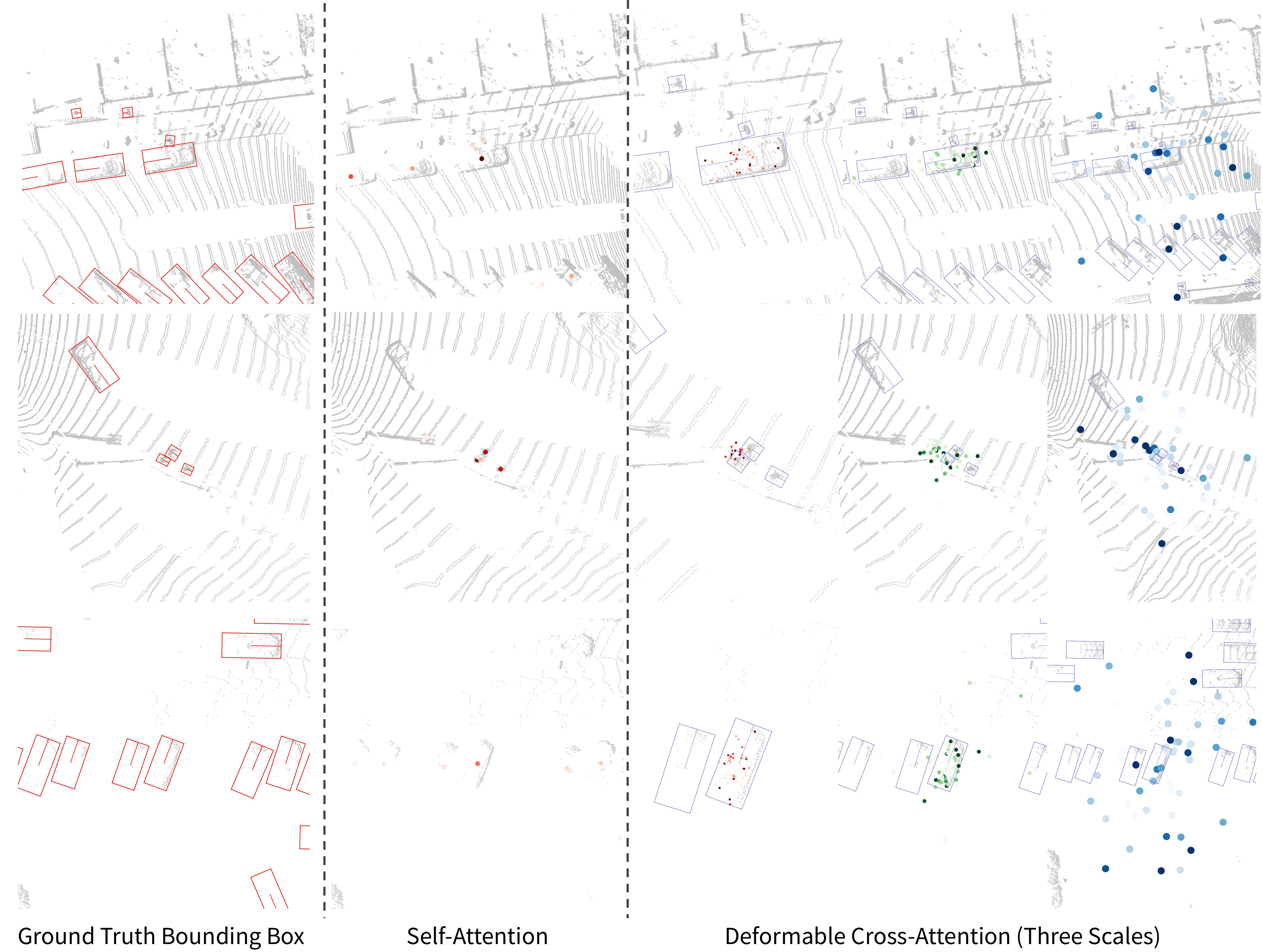}
    \caption{\textbf{The visualization of the learned self- and cross-attention weight.} The lightness or darkness of the color represents the value of attention weight. The red box denotes the ground truth box and blue box denotes the predicted box. In cross-attention, the sampled keypoints are drawn with \textcolor{red}{red}, \textcolor{green}{green} and \textcolor{blue}{blue} for the scale from low to high. Best viewed in color.}
    \label{fig:attention_vis}
    \vspace{-10pt}
\end{figure}

\noindent\textbf{Visualization of the learned attention}
The visualizations of the learned self- and cross-attention are illustrated in Figure~\ref{fig:attention_vis}. The self-attention learning is mainly focusing on the feature of the same class or nearby objects that have the same attributes. For instance, the vehicles on the same line or same parking area will have higher attention weight. The offsets learned by the deformable attention layer vary among different scales. The offsets in two lower scales mostly lead to the keypoints inside or around the object, whereas the offsets in the higher scale can sample far-range features.

\section{Conclusion}
In this paper, we propose a novel center-based transformer for 3D object detection. Our method provides a solution to improve the anchor-free 3D object detection network through object-level attention learning. Compared to the DETR-style transformer networks, we use the center feature as the initial query embedding in the transformer decoder to speed up the convergence. We also avoid high computational complexity by focusing the cross-attention learning of each query in a small multi-scale window or a deformable region. Results show that the proposed method outperforms the strong baseline in the Waymo Open Dataset, and reaches state-of-the-art performance when extended to multi-frame. We hope our design will inspire more future work in query-based transformers for LiDAR point cloud analysis.\\

\noindent\textbf{Acknowledgements} We thank Yufei Xie for his help with refactoring the code for release.

\clearpage
%
%
\bibliographystyle{splncs04}
\bibliography{egbib}

\appendix

\section{Implementation Details}

\textbf{VoxelNet backbone network}
We adopt the same VoxelNet backbone network design in CenterPoint~\cite{yin2021center}. In specific, we first use an average pooling in each voxel to encode the point cloud into a voxel feature map. Then, a VoxelNet~\cite{zhou2018voxelnet} with sparse convolution is used to extract features in the voxel map. Except for the down-sample layer, all residual blocks use the submanifold sparse convolution layer to minimize the computation cost. The VoxelNet backbone network down-samples the dimensions of the x-axis and y-axis with a factor of 8 and the z-axis with a factor of 16. Finally, the output voxel feature map is reshaped to the BEV for the following process.

\begin{figure}[t]
    \centering
    \includegraphics[width = 0.6\linewidth]{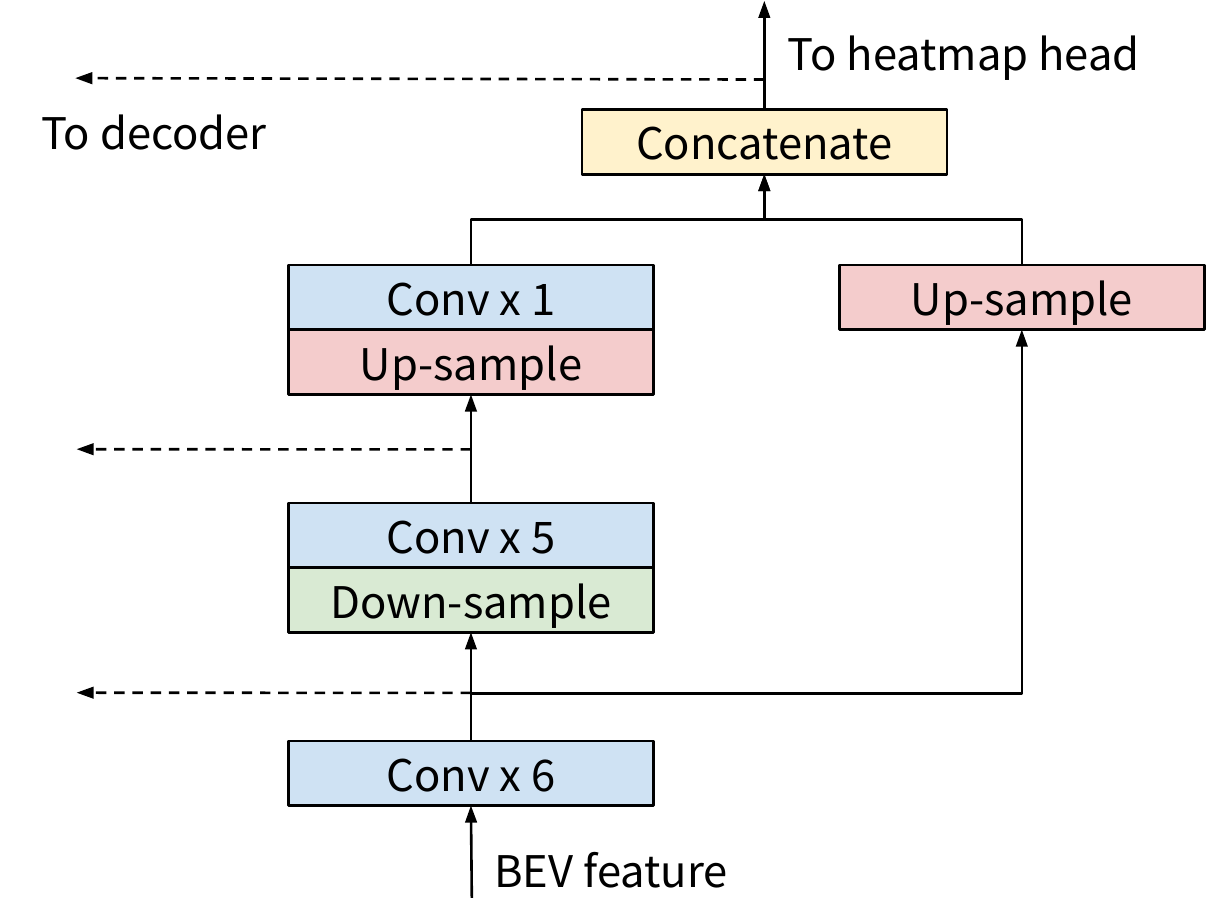}
    \caption{\textbf{The network structure of multi-scale CPN.} Each \textbf{Conv} block contains a convolution layer with kernel size $3\times3$, a batchnorm layer and a relu activation layer. We use convolution layer with stride and transpose convolution layer as the down-sample and up-sample layers.}
    \label{fig:CPN}
\end{figure}

\noindent\textbf{Multi-scale CPN}
We design the Multi-scale CPN to achieve two goals: First, we want to encode the BEV feature into different scale levels for the transformer decoder. Second, the BEV feature map should be large enough to separate small objects like the pedestrian. Since in our experiment, the size of each BEV grid in the VoxelNet output feature is [0.8m $\times$ 0.8m], which is similar to the average pedestrian object size, we need to up-sample the feature map to avoid the voxelization error. We also use a down-sample layer to extract larger-scale features. The overall network structure is shown in Figure~\ref{fig:CPN}.

\noindent\textbf{Spatial-aware heatmap fusion} To focus the heatmap fusion on the current object center location, we use the current BEV feature as the reference to learn spatial attention. We concatenate the current feature and the weighted previous features together and use another $3\times3$ convolution layer to compress the BEV feature into the same size as the single frame input in the heatmap head. 

\noindent\textbf{Training details} Generally, the transformer decoder requires a matching process, e.g. Hungarian matching, to find the closest ground truth bounding box to the prediction in training. The computation cost of this process becomes unacceptable when we match two 3D bounding boxes with orientation. Hence, we use the same training strategy in the center-based object detection network, i.e. only training the network when the proposed center is at the same position as the ground truth center. To utilize all annotation information in training, we manually select the center positions of all ground truth bounding boxes as the initial center proposals in training. And the position with the highest heatmap response other than those positions are selected as the remaining proposals. This allows the network to have a meaningful training objective from the beginning of the training, and thus converges faster. 

\section{NuScenes Dataset Result}

\textbf{NuScenes Dataset (ND)~\cite{caesar2020nuscenes}} is a large-scale dataset created by Motional. It contains 1000 scenes of 20s each, which are split into 700,150,150 sequences for the training, validation and testing. ND uses a 32 lines LiDAR with the frequency of 20 FPS. In each keyframe that is sampled every 0.5s, ND provides bounding box annotations for 10 different classes. The evaluation metrics used by ND are mean average precision (mAP) and nuScenes detection score (NDS). In contrast to WOD, the mAP used by NP only matches objects according to the 2D center distance in BEV rather than IoU. NDS is computed based on a weighted sum of \textbf{A}verage \textbf{T}ranslation/ \textbf{S}cale/ \textbf{O}rientation/ \textbf{V}elocity and \textbf{A}ttribute \textbf{E}rrors (ATE/ ASE/ AOE/ AVE/ AAE) on the set of true positives. Followed by \cite{yin2021center}, we set the range of the 3D voxel space as $[-54\text{m},54\text{m}]$ for the $X$ and $Y$ axes, and $[-5\text{m},3\text{m}]$ for the $Z$ axis. The size of each voxel is set to $(0.075\text{m},0.075\text{m},0.2\text{m})$.

\begin{table}[t]
\centering
\captionof{table}{The detection result on the ND validation set. $\ddagger$: Base CenterFormer model without IoU rectification and multi-frame fusion.  $*$ : Our implementation uses the same backbone network and training configuration.}
\label{table:nuscenes}
\resizebox{0.8\linewidth}{!}{
\begin{tabular}{c|ccccccc}
\Xhline{4\arrayrulewidth}
Method & mAP $\uparrow$ & NDS $\uparrow$ & ATE $\downarrow$ & ASE $\downarrow$ & AOE $\downarrow$ & AVE $\downarrow$ & AAE $\downarrow$  \\ \Xhline{2\arrayrulewidth}
PointPillars~\cite{lang2019pointpillars} & 39.3 & 53.3 & - & - & - & - & - \\
Pillar-OD~\cite{wang2020PillarOD} & 44.4 & 56.8 & - & - & - & - & - \\
SSN~\cite{zhu2020ssn} & 45.3 & 57.0 & - & - & - & - & - \\
CBGS~\cite{zhu2019class} & 51.4 & 62.6 & - & - & - & - & - \\
CenterPoint$^{*}$~\cite{yin2021center}    &   55.2 & 64.4 & 29.3 & 25.7 & 29.6 & 27.2 & \textbf{19.5} \\ 
\Xhline{2\arrayrulewidth}
CenterFormer$^{\ddagger}$    &   \textbf{55.4} &  \textbf{65.2} & \textbf{27.5} & \textbf{25.2} & \textbf{27.5} & \textbf{24.3} & 20.8 \\
\Xhline{4\arrayrulewidth}
\end{tabular}
}
\end{table}

We show the comparison of the results on the nuScenes validation set in Table~\ref{table:nuscenes}. We compare our base CenterFormer model with the CenterPoint baseline using the same training configuration. Due to the time limitation, we did not include further experiments on ND using some of our more complex structures, like deformable cross-attention and multi-frame fusion.
Nevertheless, our base CenterFormer already outperforms CenterPoint as shown in the table. The improvement comes mainly from the bounding box regression since these two methods share a similar center-based classification design. This result validates the superiority of our proposed CenterFormer over the traditional center-based detector in different point cloud structures.

\section{Analysis}

\begin{figure}[t]
\centering
\begin{minipage}{0.53\textwidth}
\centering
\captionsetup{type=table}
\caption{The LEVEL\_2 mAPH results comparison of the multi-frame CenterFormer on WOD validation set. All models are trained without the IoU rectification. $\star$: CenterFormer using point concatenation. $\dagger$: Deformable CenterFormer.}
\label{table:multi_frame}
\resizebox{\linewidth}{!}{
\begin{tabular}{l|c|ccc>{\columncolor[gray]{0.95}}c}
\Xhline{4\arrayrulewidth}
 Method & Frame &Vehicle & Pedestrian & Cyclist & Mean \\
\Xhline{2\arrayrulewidth}
 CenterFormer         & 1 &69.0 & 66.8 & 68.0 & 67.9 \\
\Xhline{2\arrayrulewidth}
 CenterFormer$^{\star}$ & 2 &70.6 & 70.2 & 70.7 & 70.5 \\

 CenterFormer$^{\star}$ & 4 &71.7 & 70.8 & 71.6 & 71.4 \\

 CenterFormer$^{\star}$ & 8 &72.0 & 71.6 & 71.6 & 71.7 \\
\Xhline{2\arrayrulewidth}
 CenterFormer & 2 &70.9 & 70.4 & 71.8 & 71.0 \\

 CenterFormer$^{\dagger}$ & 2 &70.7 & 71.1 & 72.6 & 71.5 \\

 CenterFormer$^{\dagger}$ & 4 &71.9 & 72.2 & 71.5 & 71.9 \\

 CenterFormer$^{\dagger}$ & 8 &73.4 & 73.4 & 71.7 & 72.8 \\
\Xhline{4\arrayrulewidth}
\end{tabular}
}
\end{minipage}
~
\begin{minipage}{0.44\textwidth}
\centering
\caption{The LEVEL\_2 mAPH results comparison breakdown by speed.}\label{fig:speed}
\resizebox{\linewidth}{!}{
\includegraphics{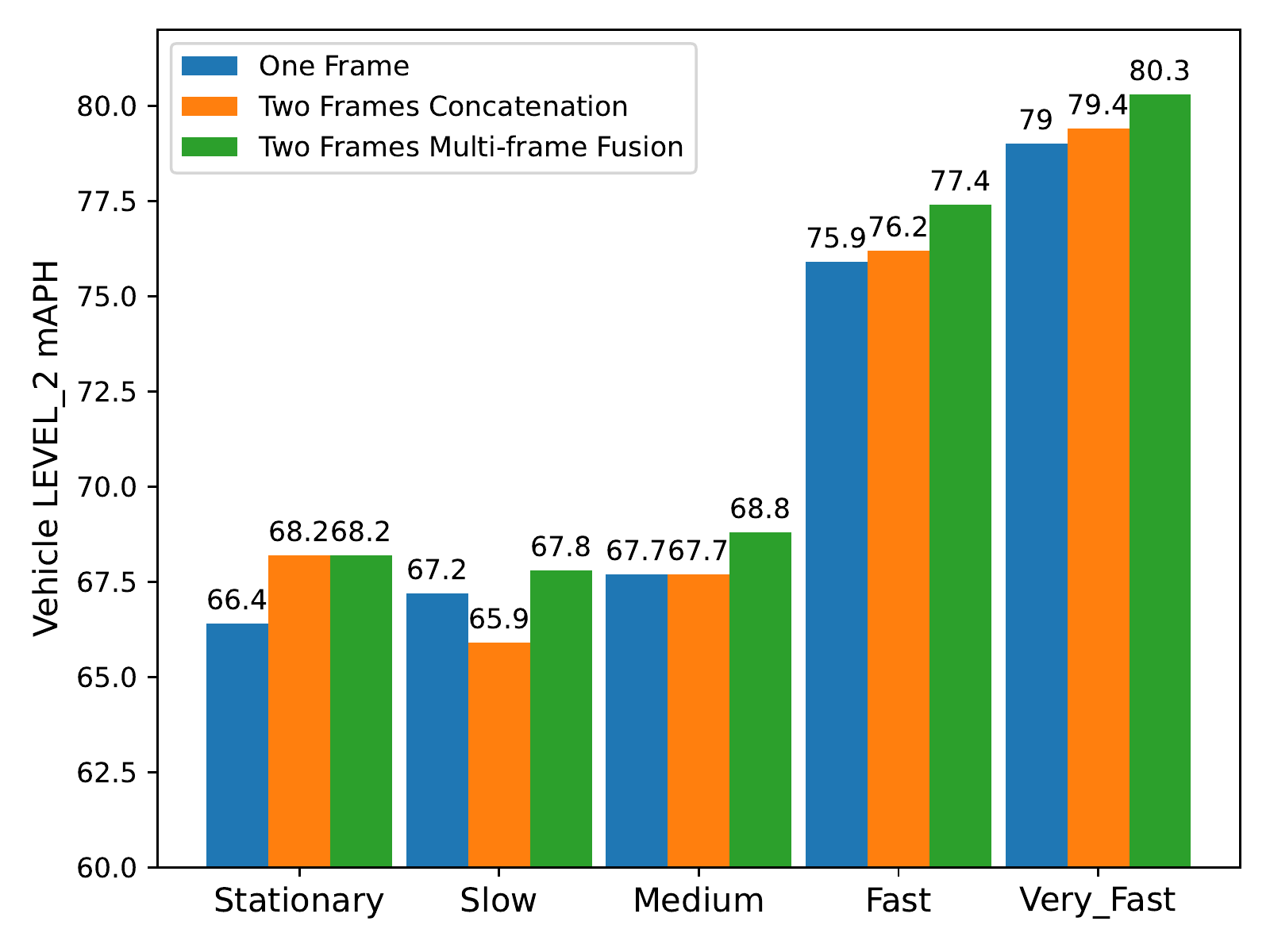}
}
\end{minipage}
\end{figure}

\textbf{The effect of our multi-frame design}
In Table~\ref{table:multi_frame}, we show the improvement of our multi-frame CenterFormer compared to the point concatenation method used by most LiDAR object detectors. The point concatenation method has significant improvement ($+2.6\%$) on two frames. But it has less effect when using more frames. In contrast, our multi-frame CenterFormer has constant improvement when using more frames. In 2, 4 and 8 frames, multi-frame CenterFormer achieves $1.0\%$, $0.5\%$ and $1.1\%$ higher mAPH than the point concatenation method. Our deformable CenterFormer achieves better performance than the base model on multi-frame due to the ability to model cross-attention in a larger range. We also compared the performance on different speeds in Figure~\ref{fig:speed}. The speed of a object is categorized into stationary ($<0.2m/s$), slow ($0.2\sim 1m/s$), medium ($1\sim 3m/s$), fast ($3\sim 10m/s$) or very fast ($>10m/s$). We can see that the main improvement in the point concatenation method comes from the stationary objects, and the slow-speed objects even have worse performance. On the contrary, our multi-frame CenterFormer achieves better performance throughout all categories.

\begin{table}[t]
\centering
\caption{The LEVEL\_2 mAPH result comparison of using different layer and head configurations on WOD validation set. (Left) Base CenterFormer. (Right) Deformable CenterFormer. \textbf{All experiments use only 20\% of uniformly sampled training data.}}
\label{table:layer_head}
\begin{minipage}{0.45\textwidth}
\resizebox{\linewidth}{!}{
\begin{tabular}{cc|ccc>{\columncolor[gray]{0.95}}c}
\Xhline{2\arrayrulewidth}
layer & head & Vehicle & Pedestrian & Cyclist & Mean \\ \Xhline{2\arrayrulewidth}
2     & 4    &   65.5      &   61.7       &   63.9      &  63.7    \\ 
2     & 6    &   65.2      &   61.0       &   64.5      &  63.7    \\ 
3     & 2    &   65.2      &   61.0       &   63.3      &  63.2    \\ 
\rowcolor{lightgray!30} 3     & 4    &   65.4      &   61.6       &   65.1      &  64.0    \\ 
3     & 6    &   65.0      &   61.6       &   64.5      &  63.7    \\ 
4     & 2    &   64.2      &   60.7       &   64.1      &  63.0    \\ 
4     & 4    &   64.9      &   61.4       &   64.1      &  63.5    \\ \Xhline{2\arrayrulewidth}
\end{tabular}
}
\end{minipage}
~~~~~~
\begin{minipage}{0.45\textwidth}
\resizebox{\linewidth}{!}{
\begin{tabular}{cc|ccc>{\columncolor[gray]{0.95}}c}
\Xhline{2\arrayrulewidth}
layer & head & Vehicle & Pedestrian & Cyclist & Mean \\ \Xhline{2\arrayrulewidth}
1     & 3    &   65.1      &   60.2        &   64.7      &  63.3    \\ 
2     & 3    &   64.9      &   60.7       &   64.8      &  63.4    \\ 
\rowcolor{lightgray!30} 2     & 6    &   65.3      &   60.4        &   66.0      &  63.9    \\ 
3     & 3    &   65.7      &   60.7        &   65.1      &  63.7    \\ 
3     & 6    &   64.5      &   60.3        &   64.5      &  63.1   \\ \Xhline{2\arrayrulewidth}
\end{tabular}
}
\end{minipage}
\end{table}

\noindent\textbf{Transformer layer and head number}
We show the comparison of results using different transformer layers and head numbers in Table~\ref{table:layer_head}. The results indicate that more transformer layers and heads do not assure better performance. The base transformer model with 3 layers and 4 heads and the deformable transformer model with 2 layers and 6 heads has the best performance.

\begin{table}[t]
\centering
\caption{The LEVEL\_2 mAPH result comparison of using different window sizes in our base CenterFormer and different offset numbers in deformable CenterFormer on  WOD validation set. (Left) Base CenterFormer. (Right) Deformable CenterFormer. \textbf{All experiments use only 20\% of uniformly sampled training data.}}
\label{table:attention_field}
\begin{minipage}{0.47\textwidth}
\resizebox{\linewidth}{!}{
\begin{tabular}{c|ccc>{\columncolor[gray]{0.95}}c}
\Xhline{2\arrayrulewidth}
Window size & Vehicle & Pedestrian & Cyclist & Mean \\ \Xhline{2\arrayrulewidth}
\rowcolor{lightgray!30} 3,3,3    &   65.4      &   61.6       &   65.1      &  64.0    \\ 
5,3,3    &   65.4      &   61.7       &   65.0      &  64.0    \\ 
5,5,3   &   65.4      &   61.8       &   64.1      &  63.8    \\ 
5,5,5   &   64.1      &   60.1       &   63.8      &  62.7   \\ 
7,3,3    &    64.9      &   61.1       &   64.7      &  63.6   \\  \Xhline{2\arrayrulewidth}
\end{tabular}
}
\end{minipage}
~
\begin{minipage}{0.47\textwidth}
\resizebox{\linewidth}{!}{
\begin{tabular}{c|ccc>{\columncolor[gray]{0.95}}c}
\Xhline{2\arrayrulewidth}
Offset number & Vehicle & Pedestrian & Cyclist & Mean \\ \Xhline{2\arrayrulewidth}
5    &   64.8      &   59.5        &   64.2      &  62.8    \\ 
9    &   64.4      &   60.6      &   64.5      &  63.2    \\ 
\rowcolor{lightgray!30} 15    &   65.3      &   60.4        &   66.0      &  63.9    \\ 
20    &   65.0      &   60.2        &   64.3      &  63.2    \\  \Xhline{2\arrayrulewidth}
\end{tabular}
}
\end{minipage}
\end{table}

\noindent\textbf{Cross-attention field}
We show the comparison of results using different cross-attention window sizes and offset numbers in Table~\ref{table:attention_field}. Increasing the window size does not have any performance gain. This is because the sizes of each grid in three scales are $[0.4m, 0.8m, 1.6m]$ in our setting. The $3\times3$ attention window can encompass the region of almost all pedestrian and cyclist objects. If we increase it to $5\times5$ or $7\times7$, although it can include more features for the vehicle, the added feature for the pedestrian and cyclist is almost unrelated. On the other hand, the number of offsets used in the deformable cross-attention also does not increase the performance monotonically. We find the offset number of 15 has the best performance.

\begin{figure}[t]
\centering
\begin{minipage}{0.47\textwidth}
\centering
\captionsetup{type=table}
\caption{The LEVEL\_2 mAPH result comparison of the position encoding methods on WOD validation set. \textbf{All experiments use only 20\% of uniformly sampled training data.}}
\label{table:position_encoding}
\resizebox{\linewidth}{!}{
\begin{tabular}{l|ccc>{\columncolor[gray]{0.95}}c}
\Xhline{4\arrayrulewidth}
 Encoder & Vehicle & Pedestrian & Cyclist & Mean \\
\Xhline{2\arrayrulewidth}
 None & 60.3 & 56.3 & 61.3 & 59.5 \\
\hline 
 Sinusoidal & 62.7 & 58.6 & 63.5 & 61.7 \\
\hline
 Linear & 65.2 & 60.9 & 66.0 & 64.0 \\
\Xhline{4\arrayrulewidth}
\end{tabular}
}
\end{minipage}
~
\begin{minipage}{0.42\textwidth}
\centering
\caption{The LEVEL\_2 mAPH results comparison of CenterFormer and DETR in different epochs.}
\label{pic:plot_epoch}
\resizebox{\linewidth}{!}{
\includegraphics{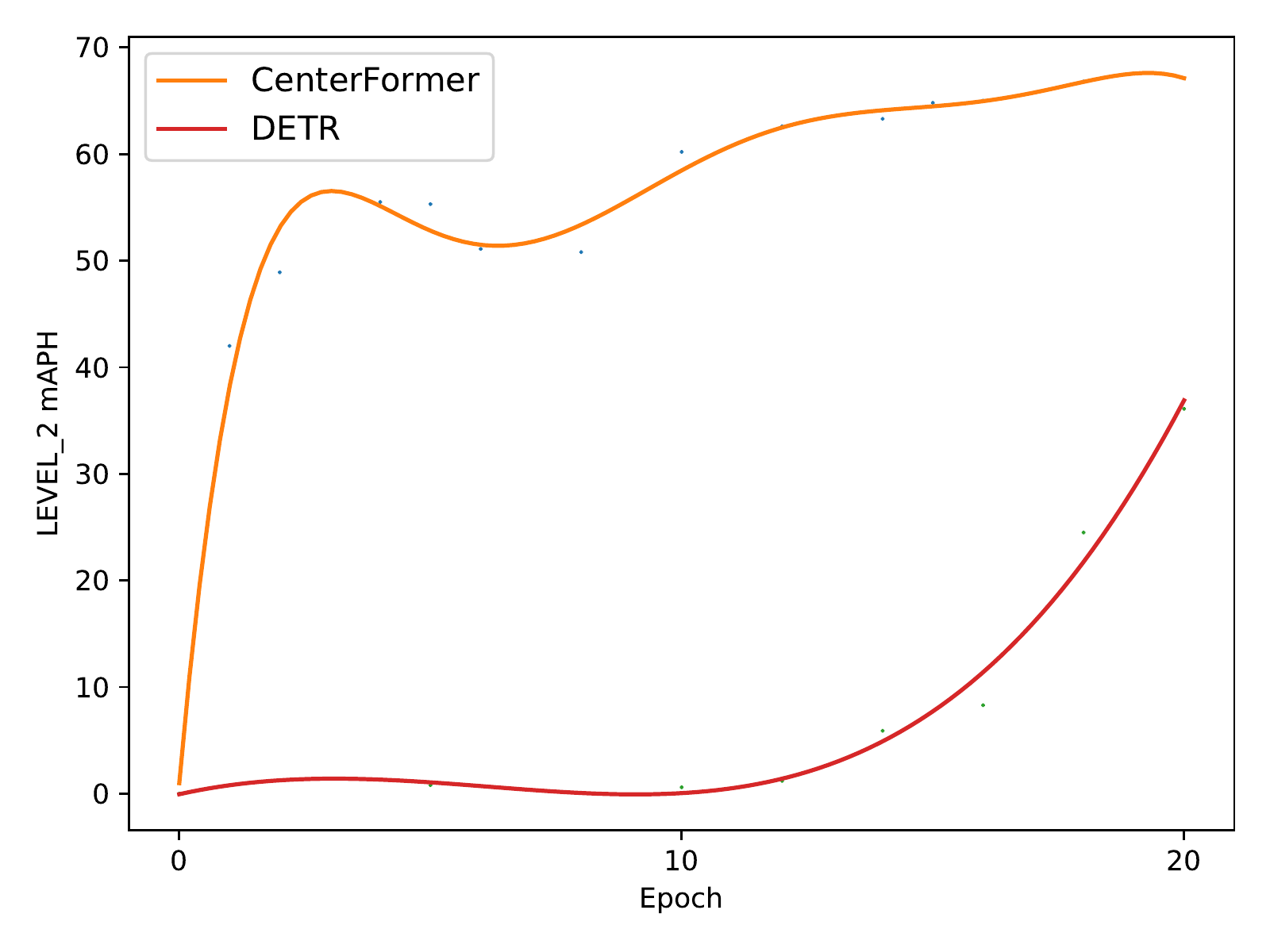}
}
\end{minipage}
\end{figure}

\noindent\textbf{Position embedding}
Position embedding is important in the transformer model to capture the spatial relationship between input features. Standard position embedding is either crafted manually using sine and cosine distances or learned through a linear layer. However, 3D point clouds, as a specific type of data, contain the position information in the raw feature. They do not necessarily need the position embedding since the spatial information is already in the encoded feature. We test the performance of different position embedding methods using 20\% of training data. The LEVEL\_2 mAPH result is shown in Table~\ref{table:position_encoding}. We can see without the position embedding, the result drops significantly to 59.5\%. This indicates the absolute position information is still an important feature to guide the attention learning of the transformer model. The learnable encoding method also outperforms the sinusoidal encoding method by a large margin. 

\noindent\textbf{Converging difficulty compared with DETR}
In Figure~\ref{pic:plot_epoch}, We show the LEVEL\_2 mAPH result comparison of CenterFormer and DETR in a 20 epochs training cycle. We implement the DETR-style set matching training strategy using the same backbone network in CenterFormer. We can see that not only CenterFormer can reach a much higher result than DETR, but also converge much faster than DETR.

\begin{figure}[t!]
    \centering
    \includegraphics[width = \linewidth]{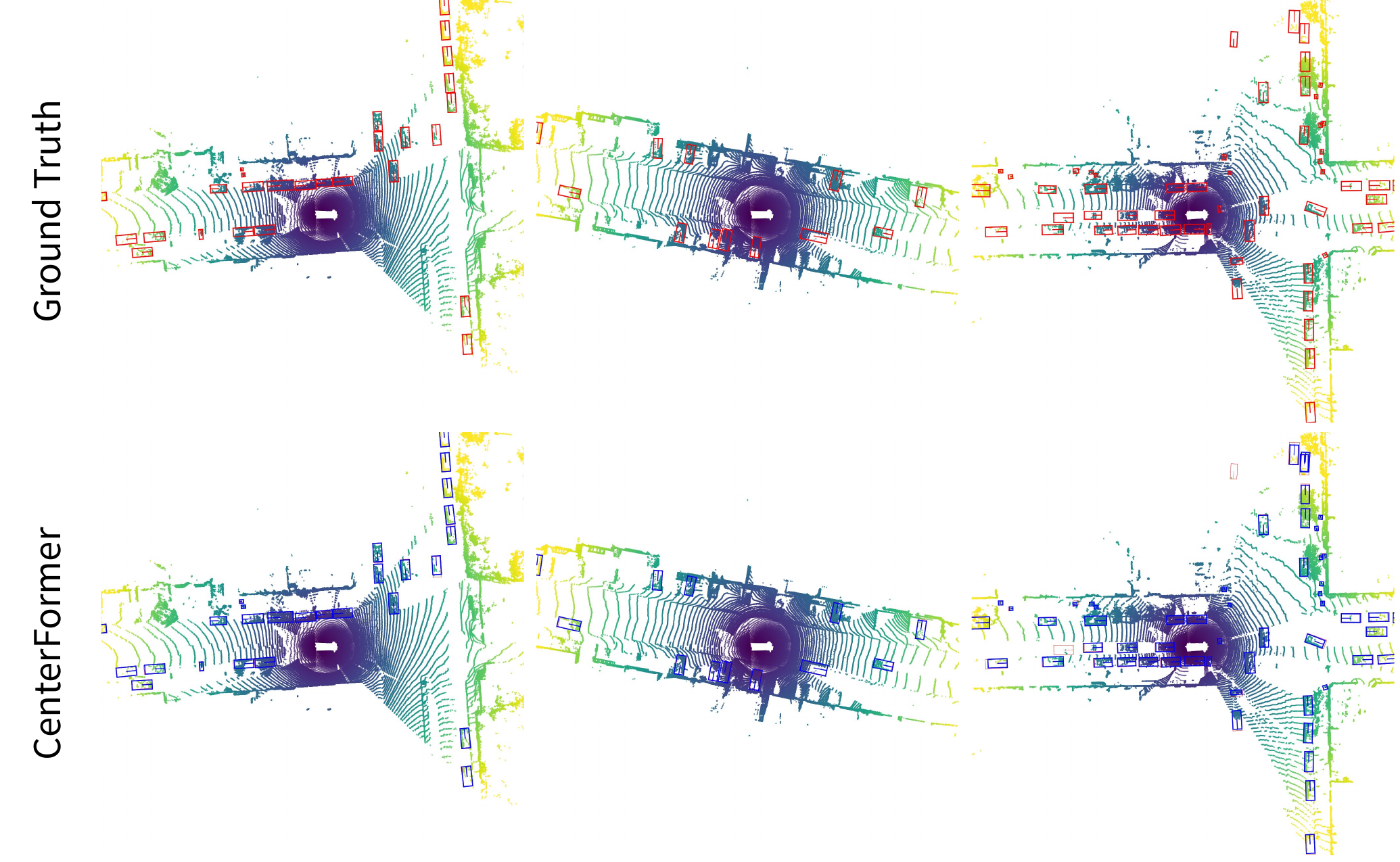}
    \caption{\textbf{Visualization of CenterFormer predictions.} The red box denotes the ground truth bounding box. The blue box denotes the predictions with confidence score $>0.4$. Truncated objects whose center is outside of the 50m range are not visualized. Best viewed in color.}
    \label{fig:qualitative}
\end{figure}

\noindent\textbf{Comparison with two-stage LiDAR detection}
Most two-stage LiDAR detection methods~\cite{shi2019pointrcnn,shi2020pv,yin2021center,li2021lidar} apply the RCNN-style refinement network in the 3D domain. The second stage aggregates RoI features in each first-stage proposal to refine both the classification and regression prediction. There are two drawbacks in this design. First, the second stage only utilizes local RoI features. It cannot retrieve global information and depends heavily on the quality of the first feature and proposal. Second, it has a large computation overhead, especially when used in LiDAR point clouds with a large size of points or voxel features. The network will predict the same object information twice, which results in a cumbersome structure and prohibitive run-time. In contrast, our method works between one-stage and two-stage. We use a center proposal network to generate initial center queries. The self-attention layer allows the network to directly learn object-level contextual information. The cross-attention layer can also capture long-range information in the multi-scale BEV feature. The classification and regression are done only once in our method.

\section{Qualitative Results}
Figure~\ref{fig:qualitative} shows the qualitative result of our proposed method. Our method can make accurate predictions with a high confidence score.

\end{document}